\documentclass[times,twocolumn,final,authoryear]{elsarticle}

\usepackage{yiswa}
\usepackage{framed,multirow}

\usepackage{amsmath}
\usepackage{mathtools}
\usepackage{amssymb}
\usepackage[ruled,vlined]{algorithm2e}
\usepackage{tabularx}
\usepackage{booktabs}
\usepackage{caption}
\usepackage{geometry}

\usepackage{amssymb}
\usepackage{latexsym}

\usepackage{url}
\usepackage{xcolor}
\definecolor{newcolor}{rgb}{.8,.349,.1}
\usepackage[colorlinks,citecolor=red,urlcolor=blue,bookmarks=false,hypertexnames=true]{hyperref}

\journal{Intelligent Systems with Applications}

\begin{document}

\ifpreprint
  \setcounter{page}{1}
\else
  \setcounter{page}{1}
\fi

\begin{frontmatter}

\title{Facial Expression Video Generation Based-On Spatio-temporal Convolutional  GAN: FEV-GAN}

\author[1]{Hamza \snm{Bouzid} \corref{cor1}} 
\cortext[cor1]{Corresponding author
  }
\ead{hamza-bouzid@um5r.ac.ma }
\author[1]{Lahoucine \snm{Ballihi} }
\ead{lahoucine.ballihi@fsr.um5.ac.ma}

\address[1]{LRIT-CNRST URAC 29, Mohammed V University in Rabat, Faculty Of Sciences, Rabat, Morocco.}

\received{}
\finalform{}
\accepted{}
\availableonline{}
\communicated{}

\begin{abstract}Facial expression generation has always been an intriguing task for scientists and researchers all over the globe. In this context, we present our novel approach for generating videos of the six basic facial expressions. Starting from a single neutral facial image and a label indicating the desired facial expression, we aim to synthesize a video of the given identity performing  the specified facial expression.  Our approach, referred to as FEV-GAN (Facial Expression Video GAN), is based on Spatio-temporal Convolutional GANs, that are known to model both content and motion in the same network. Previous methods based on such a network have shown a good ability to generate coherent videos with smooth temporal evolution. However, they still suffer from low image quality and low identity preservation capability. In this work, we address this problem by using a generator composed of two image encoders. The first one is pre-trained for facial identity feature extraction and the second for spatial feature extraction. We have qualitatively and quantitatively evaluated our model on two international facial expression benchmark databases: MUG and Oulu-CASIA NIR\&VIS. The experimental results analysis demonstrates the effectiveness of our approach in generating videos of the six basic facial expressions while preserving the input identity. The analysis also proves that the use of both identity and spatial features enhances the decoder ability to better preserve the identity and generate high-quality videos. The code and the pre-trained model will soon be made publicly available.
\end{abstract}

\begin{keyword}
\MSC 41A05\sep 41A10\sep 65D05\sep 65D17
\KWD Facial expression video generation \sep Deep learning \sep Generative adversarial networks \sep Spatio-temporal convolutional networks
\end{keyword}

\end{frontmatter}

%\linenumbers

\section{Introduction}
\label{introduction}

Facial expressions have always been considered one of the essential tools for human interaction. Integrating the ability to recognize and synthesize facial expressions to machines provides a natural and smooth interaction. Which opens the door to many exciting new applications in different fields, including the movie industry, e-commerce, and even in the medical field. Motivated by this, researches have studied facial expression recognition and have already reached a high level of precision, while facial expression generation has been more demanding and less studied in the state of the art. Recently, with the success of Generative Adversarial Networks (GANs) \citep{goodfellow2014generative} in data generation, in particular image generation, the task of generating facial expressions has seen tremendous progress.

However, the dynamic facial expressions synthesis is even less studied due to the difficulty of the tasks: \textbf{1)} learning the dataset distribution (facial structure, background), \textbf{2)} representing a natural and smooth evolution of facial expressions (Temporal representation), and \textbf{3)} preserving the same input identity. To address the high complexity of these three tasks, most existing methods tend to treat facial expression generation as a two-steps process. One step for the low dimensional temporal generation (motion) and the other for the spatial generation (content). Such methods \citep{tulyakov2018mocogan, wang2018every, otberdout2019dynamic} are mostly based on \textbf{1)} the generation of motion as codes in a latent space, thereafter \textbf{2)} combining it with the input image embedding to generate frames individually, through the use of an Image-to-Image translation network. These methods are efficient in learning facial structure and identity preservation, but they are flawed when it comes to modeling spatio-temporal consistency and appearance retainment. This is caused by the independence between frames generation.\\

Motivated by the success of Deep spatio-temporal neural network models in recognition and prediction tasks \citep{al2018deep,tran2018closer,ali2021exploiting,ali2022exploiting}, researchers have proposed a variety of one-step methods \citep{vondrick2016generating,jang2018video,wang2020g3an,wang2020imaginator} that use fractionally strided 3D convolutions. Videos generated by these kinds of methods show more spatio-temporal consistency, however, lower video quality, more noise and distortions, and more identity preservation issues compared to two-steps methods.  We claim this is due to the high complexity of the three tasks combined in a single network ( learning \textbf{1.} the spatial presentation, \textbf{2.} the temporal representation, \textbf{3.} identity preservation). This requires a large network with high potential complexity and a large amount of data, which significantly increases the difficulty of model optimization.\\

To solve the issues of the low quality, noise and identity preservation capability faced by one-step methods,  We propose encoding the input image into two codes in the latent space, using two feature extractors ($E_{Id}$ identity feature extractor, $E_s$ spatial feature extractor). We also suggest exploiting the high performance of state-of-the-art facial recognition systems, by utilizing a pre-trained facial recognition feature extractor as our identity encoder $E_{Id}$. This grants identity related features that help with identity preservation. In addition, the use of a pre-trained feature extractor allows for applying the optimization process only on the other encoder $E_S$, that is used to extract other spatial features in order to maintain sufficiently good quality while reconstructing facial expression video. \\

In summary, our contributions include the following aspects:
\begin{enumerate}
    \item We propose a conditional GAN, with a single generator and a single discriminator, that generates at each time step a dynamic facial expression video, corresponding to the desired class of expressions. The generated videos present a realistic appearance, and preserve the identity of the input image.
    \item We investigate the influence of utilizing two encoders {$E_{Id}$ and $E_{S}$}, where $E_{Id}$ is a facial identity feature extractor and $E_{S}$ is a spatial feature extractor. 
    \item We exploit the high potential of state-of-the-art facial recognition systems. We use a pre-trained face recognition model as our generator encoder $E_{Id}$, which will ensure strongly related identity features. This aims to facilitate the task of the decoder by providing meaningful and structured features. 
    \item We deeply evaluate our model, quantitatively and qualitatively, on two public facial expressions benchmarks: MUG facial expression database and Oulu-CASIA NIR\&VIS facial expression database. We compare it with the recent state-of-the-art approaches: VGAN \citep{vondrick2016generating}, MoCoGAN \citep{tulyakov2018mocogan}, ImaGINator \citep{wang2020imaginator} and \citep{otberdout2019dynamic}.
\end{enumerate}

\section{Related Work}

\textbf{Static Facial Expression Generation $-$}
Facial expressions synthesis was initially achieved through the use of traditional methods, such as geometric interpolation \citep{pighin2006synthesizing},Parameterization \citep{raouzaiou2002parameterized}, Morphing \citep{beier1992feature}, etc. These methods show success on avatars, but they fall short when dealing with real faces, and they are unable to generalize a flow of movement for all human faces due to the high complexity of natural human expressions and the variety of identity-specific characteristics. To face these limitations, neural networks based methods have been applied on facial expressions generation, including RBMs \citep{zeiler2011facial}, DBNs \citep{sabzevari2010fast} and Variational Auto-Encoders \citep{kingma2013auto},etc. These methods learn acceptable data representations and better flow between different data distributions compared to prior methods, but they face problems such as the lack of precision in controlling facial expressions, low resolution and blurry generated images.

With the appearance of GANs, multiple of its extensions have been dedicated to facial expressions generation. \citep{makhzani2015adversarial} and \citep{zhou2017photorealistic} exploit the concept of adversity with auto-encoders to present Adversarial Auto Encoders.
\citep{zhu2017unpaired} propose a conditional GAN, namely CycleGAN, that uses Cycle-Consistency Loss to preserve the key attributes (identity) of the data.  
\citep{choi2018stargan} address the inefficiency of creating a new generator for each type of transformation, by proposing an architecture that can handle different transformation between different datasets.
\citep{wang2018facial} suggest exploiting the U-Net architecture as GAN generator, in order to increase the quality and the resolution of generated images. 
US-GAN \citep{akram2021us} uses a skip connection, called the ultimate skip connection, that links the input image with the output of the model, which allows the model to focus on learning expression-related details only. the model outputs the addition of the input image and the generated expression details, improving therefore the identity preservation, but displaying artifacts in areas related to the expression (mouth, nose, eyes).  
The studies above established the task of generating classes of facial expressions (sad, happy, angry, etc.), but in reality, the intensity of the expression inhibits the understanding of the emotional state of the person. 
ExprGAN \citep{ding2017exprgan} used an expression controller module to control the expression intensity continuously from weak to strong. Methods like GANimation \citep{pumarola2018ganimation}, EF-GAN \citep{wu2020cascade} used Action Units (AUs) in order to learn conditioning the generation process which offers more diversity in the generated expressions. Other methods such as G2-GAN \citep{song2018geometry} and GC-GAN \citep{qiao2018geometry} exploited Facial Geometry as a condition to control the facial expression synthesis. The objective of the latter models is to take as input a facial image and facial Landmarks in form of binary images or landmarks coordinates, then learn to generate a realistic face image with the same identity and the target expression. \citep{kollias2020deep} and \citep{bai2022data} utilize labels from the 2D Valence-Arousal Space, in which the valence is how positive or negative is an emotion and the arousal is the power of the emotion activation \citep{russell1980circumplex}, to guide the process of facial expression generation, enhancing the variety and the control of the generated expressions. All these approaches and others have established the task of facial expression generation, but have not considered the dynamicity of these expressions. \\

\textbf{Dynamic Facial Expression Generation $-$ }  Facial expressions are naturally dynamic actions that contain more information and details than a single pose, e.g. the speed of facial expression transformation, head movements when displaying the expression, etc. This information can be significant in understanding the emotional state of a person. 
To achieve this, methods like \citep{ha2020marionette, tang2021eggan, li2021facial, tu2021image,vowels2021vdsm} focus on facial expression transfer, in which the facial expression is transferred from a driver to a target face, while aiming to preserve the target identity even in situations where the facial characteristics of the driver differs widely from those of the target.
In other methods, the motion is generated separately as codes in the latent space, these codes are then fed to the generator in order to generate frames of the video individually. For example,  MoCoGAN \citep{tulyakov2018mocogan} decomposes the video into content and motion information, where the video motion is learned by Gated RNN (GRU) and the video frames are generated sequentially by a GAN. RV-GAN \citep{gupta2022rv} uses a transpose (upsampling instead of downsampling) convolutional LSTMs as GAN generator to generate frames individually. However, the results of both models present content and motion artifacts, and they both could only be applied to seen-before identities, and a finite number of expressions. In \citep{fan2019controllable}, the principle of MoCoGAN is extended by adding an encoder that helps preserving the input identity, and a coefficient to control the degree of the expression continuously. The authors of \citep{wang2018every} utilize a Multi-Mode Recurrent Landmark Generator to learn generating variant sequences of facial landmarks of the same category (e.g. different ways to smile), translated later to video frames. In \citep{otberdout2019dynamic}, the authors exploit a conditional version of manifold-valued Wasserstein GAN to model the facial landmarks motion as curves encoded as points on a hypersphere. The W-GAN learns the distribution of facial expression dynamics of different classes, from which new facial expression motions are synthesized and transformed to videos by TextureGAN. 
Other works  have investigated guiding facial expression generation by speech audio data, such as \citep{chen2020talking, guo2021ad, wang2022one, liang2022expressive}, or by a combination of audio and facial landmark information, like \citep{wang2021audio2head, wu2021imitating, sinha2022emotion} . All the methods mentioned before are methods  that generate a single frame at a time-step, which lowers the dependency between the video frames causing the lack of spatio-temporal consistency.
Contrasted to the methods mentioned before, methods like VGAN\citep{vondrick2016generating}, G3AN\citep{wang2020g3an} and ImaGINator \citep{wang2020imaginator} use a single step for the generation of the whole facial expression video, by employing fractionally strided spatio-temporal convolutions to simultaneously generate both appearance, as well as motion. VGAN decomposes the generated videos into two parts, a static section (the background) and a dynamic section, which imposes the use of a generator composed of two streams, for generating the background and the foreground, that are combined in the output to generate the whole video. 
G3AN aims to model appearance and motion in disentangled manner. This is accomplished by decomposing appearance and motion in a three-stream Generator, where the main stream models spatio-temporal consistency, while the two auxiliary streams enhance the main stream with multi-scale appearance and motion features, respectively. Both VGAN and G3AN are unconditional models that start from a Gaussian noise input, causing the lack of identity preservation and of control over the generated expression.
In order to avoid these problems, ImaGINator uses a blend of auto-encoder architecture and spatio-temporal fusion mechanism, where the low-level spatial features in the encoder are sent directly to the decoder (the same concept as U-Net \citep{ronneberger2015u}). It also uses two discriminators, one processes the whole video and the other processes frame by frame. Videos generated by these kinds of methods show more spatio-temporal consistency but lower video quality, more noise and less identity preservation compared to two-steps methods.

Motivated by the discussed above, we present a novel one-step approach for facial expression generation based on fractionally strided spatio-temporal convolutions. The rest of the paper is organized as follows. In section 3 we introduce our new FEV-GAN model. Section 4 shows the experimental settings and the quantitative and qualitative analysis of the model. Section 5 concludes the paper and provides perspectives for future research.

\begin{figure*}[ht]
\begin{center}
\includegraphics[scale=0.7]{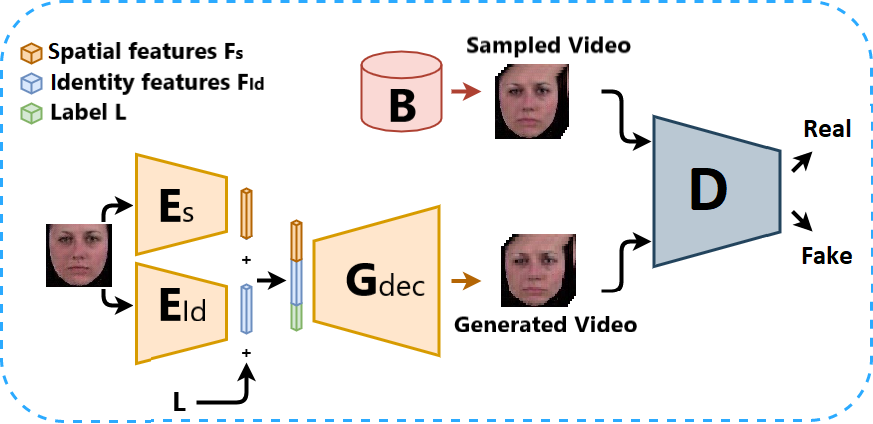}
\caption{Overview of the proposed model for generating facial expressions: \textbf{FEV-GAN}. The input image is mapped through two encoders, identity ($E_{Id}$) and spatial ($E_{S}$) encoders, into the latent codes $F_{Id}$ and $F_S$. Both codes and the label $L$ are fed to the decoder $G_{dec}$ that transforms them to a video of the input identity performing the desired expression. While the discriminator $D$ has the purpose of classifying realistic scenes from synthesized scenes.}
\label{ourApproach}
\end{center}
\end{figure*}

\section{Proposed Approach}

As stated in the introduction, our main aim is to establish a model that generates dynamic facial expression videos from appearance information and expression category. Thus, we formulate our goal as learning a function $G: \{I, L \} \longrightarrow \hat{Y}$, where $I$ is the input image, $L$ is the label vector and $\hat{Y}$ is is the generated video.

To achieve this objective, we propose a Framework consisting of the following components: a Generator network $G$ built on an encoder-decoder architecture. The encoders $E_{Id}$ and $E_{S}$ take as input a single image $I$ and extract identity features $F_{Id}$ and spatial features $F_S$. The decoder $G_{dec}$ utilizes the extracted features ($F_{Id}$,$F_S$) and a label $L$ to generate a realistic video $\hat{Y}$. Finally a discriminator $D$ assists the learning of the Generator for both appearance and expression category. The overview of our approach is shown in Fig.~ \ref{ourApproach}.

\subsection{FEV-GAN Model Description}

In the following, the architecture of our network is described, and details on the generator $G$ and the discriminator $D$ are provided. \\

\textbf{Generator $E_{Id}$, $E_{S}$, $ G_{dec}$}:
As shown in Fig.~ \ref{ourApproach}, our generator consists of three networks, a pre-trained image identity encoder $E_{Id}$, a randomly initialized encoder $E_{S}$ and a video decoder $G_{dec}$. The encoder $E_{Id}$ is a well known state-of-the-art facial recognition model VGG-FACE \citep{parkhi2015deep} feature extractor. It takes a ($64\times64\times3$) RGB image $I$ as input and transforms it into 1024 feature maps of ($14\times14$), containing the facial identity features $F_{Id}$. The encoder $E_{S}$ is initialized with a Gaussian noise. It extracts 512 feature maps of ($14\times14$) features $F_S$ that contain more spatial details beside the identity features. Then, the features $F_{Id}$, $F_S$ and the label vector $L$ are concatenated and utilized by the decoder $G_{dec}$ to generate the new video. The decoder combines spatio-temporal convolutions and fractionally strided convolutions to transform the input tensor into the high dimensional generated video. Tri-dimensional convolution offers spatial and temporal invariance. Fractional convolution is an efficient over-sampling tool, allowing to transform the latent tensor into a ($32\times64\times64\times3$) video. \\

\textbf{Discriminator $D$}: The objectives of the discriminator is to learn the ability $(i)$ to classify realistic scenes from synthesized scenes, $(ii)$ to recognize realistic motion, and $(iii)$ to detect the difference between the different classes of motion. To achieve these three tasks we use a five-layer spatio-temporal convolutional network for the purpose of learning both visual content and motion modeling. The network takes ($32\times64\times64\times3$) videos $Y$ from the dataset or $\hat{Y}$ generated by the generator $G$ and the labels vectors as input. It then checks if the video is realistic and both the motion and the label are convenient. The discriminator architecture is designed to almost invert $G_{dec}$ by replacing fractionally strided spatio-temporal convolutions with direct spatio-temporal convolutions (to sub-sample instead of over-sample), and adjust the latter layer to produce a binary real or fake classification.

\subsection{Loss Function}
To train our model $G: \{I, L\} \longrightarrow \hat{Y} $, both network $D$ and $G$ are optimized using our objective function Eq \ref{eq:FullLossFunction}.

\begin{equation}
    \mathcal{L}_ {total} (G , D) = \mathcal{L}_{adv} (G, D) + \lambda_1 \mathcal{L}_{rec} (G) + \lambda_2 \mathcal{L}_{id} (G),
\label{eq:FullLossFunction}
\end{equation}

\noindent which consists of three losses: Adversarial loss $ \mathcal{L}_{adv} $ that helps the generator learn the database distribution. Reconstruction loss $ \mathcal{L}_{rec} $ that captures the overall structure of the video and improves the quality. Identity loss $ \mathcal{L}_{id} $ which ensures facial identity details preservation. Due to the large difference we found between the three losses values, $ \lambda_1 $ and $ \lambda_2 $ parameters are used to help stabilize the training and balance the optimization of our model.

We therefore aim to solve

\begin{equation}
    G^* = arg \ \underset{G}{min} \  \underset{D}{max} \ \ \mathcal{L}_ {total}.
\label{eq:ModelOptimization}    
\end{equation}

We note that the parameters of $E_{Id}$ are frozen in the training phase, as it is a pre-trained model that already extracts facial features.\\

\textbf{Adversarial Loss:} Our conditional adversarial loss is a cross entropy loss, that is applied on both $G$ and $D$, with the aim that $G (I, L)$ learns to generate videos $\hat{Y}$ that look similar to real videos $Y$, and $D$ learns to distinguish between real samples $Y$ and generated samples $\hat{Y}$. We train the model so as $G$ aims to minimize the function while $D$ aims to maximize it.
\begin{multline}
    \underset{G}{min} \  \underset{D}{max} \ \ \mathcal{L}_{adv} = \underset{G}{min} \  \underset{D}{max} \ \  \mathbb{E}_{Y \sim P_{data}(Y)} \ [log D( \ Y\ ; \ L)] \ \\ + \  \mathbb{E}_{z \sim P_{z}(z)}  \ [1 - log D( \ G ( \ I\ ; L) \ ; \ L)].
\label{eq:AdvLoss}
\end{multline}
\vspace{0.2cm}
\textbf{Reconstruction Loss:} Our reconstruction loss at the video level is defined by

\begin{equation}
    \mathcal{L}_{rec}\ =\  [\ \vert \vert Y\ -\ \hat{Y}\ \vert \vert _{1}\ ],
\label{eq:recLoss}
\end{equation}

\noindent with the purpose of capturing the overall structure, video consistency and helping preserve the identity details. This loss is a $L_1$ norm loss between the generated videos $\hat{Y}$ and the ground truth videos $Y$. By combining this loss with $ \mathcal{L}_{adv} $, it helps the generator $G$ create more realistic videos and reconstruct smooth expression motion.\\

\textbf{Identity Loss:} 
The Identity Loss is used for identity preservation. It is a $L_1$ norm loss similar to the $\mathcal{L}_{rec}$, but while $\mathcal{L}_{rec}$ aims to minimize the $L1$ distance between pixel values of the generated and the ground truth videos, $\mathcal{L}_{id}$ aims to minimize the $L1$ distance between the identity features of the input image and the frames of generated video. We exploit the same architecture of our VGG-FACE encoder to extract the identity features from both input and output data. This loss is formalized as:
\begin{equation}
    \mathcal{L}_{id}\ = \sum _{i=0}^{N}\  [\ \vert \vert F_{vgg}(I)\ -\ F_{vgg}^i(\hat{Y})\ \vert \vert _{1}\ ],
\label{eq:IdLoss}
\end{equation}

\noindent where $F_{vgg}(I)$ are the identity features of the input image $I$, and $F_{vgg}^i(\hat{Y})$ are the identity features of the $i^{th}$ frame of $\hat{Y}$.

We furthermore use $\mathcal{L}_{id}$ on 4 frames of the video instead of 32 frames. We count on spatio-temporal consistency of the 3D convolutions to generalize the identity preservation over the rest of the video. 

\subsection{training Algorithm}
Algorithm \ref{ourAlgo} highlights the training process of the proposed FEV-GAN model. The training dataset $(I, L, Y)$ is fed into Algorithm \ref{ourAlgo}. First, the model parameters are initialized. The parameters of $E_{Id}$ are loaded from the pre-trained VGG-FACE model, while the rest of the parameters are initialized from a Gaussian distribution \textbf{(1;2)}. The outer for loops are used to learn from the data for a given number of epoch and iterations in each epoch \textbf{(3;4)}. The input image $I_i$ is fed into the encoders $E_{Id}$ \textbf{(5)} and $E_{s}$ \textbf{(6)}, that encode it to identity $F_{Id}$ and spatial $F_{s}$ features, respectively. These features and the given label $L_i$ are then used by the decoder $G_{dec}$ to generate a fake video $\hat{Y_i}$ \textbf{(7)}. Next, the discriminator network $D$ estimates the probability that a video is sampled from the dataset rather than the generator \textbf{(8;9)}. The generated videos and the discriminator estimations are used to calculate the losses \textbf{(10;12;13;14)}. Finally, the back propagation method and Adam optimizer are used to train the FEV-GAN model \textbf{(11;15)}. 

\begin{algorithm}[ht]
\SetAlgoLined
\textbf{Input}: Ground-truth Video $Y = [Y_0,Y_1,...,Y_n]$; Input Image $I  =[I_0,I_1,...,I_n]$; Target expression label $L = [L_0,L_1,...,L_n]$;\newline
\textbf{Output}: FEV-GAN model; \newline
\nl $E_{Id} \ \longleftarrow \ $ initialized with non-learnable pre-trained VGG-FACE parameters;\newline
\nl $E_{s}, G_{dec}, D \ \longleftarrow \ $ initialized with learnable parameters sampled from a Gaussian distribution;\newline
\nl \For{ the number of epochs} {
    \nl \For{the number of iterations in an epoch}{
        \nl  $F_{Id}(I_i) \ \longleftarrow \ E_{Id} (I_i) $;\newline
        \nl $F_{s}(I_i) \ \longleftarrow \ E_{s} (I_i) $; \newline 
        \nl $\hat{Y_i} \ \longleftarrow \ G_{dec} (F_{Id}(I_i), F_{s}(I_i), L_i)$;\newline
        \nl $E_{real} \ \longleftarrow \ D(Y_i,L_i)$; \newline
        \nl $E_{recon} \ \longleftarrow \ D(\hat{Y_i}, L_i)$;\newline
        \nl $\mathcal{L}_{D} \ \longleftarrow \log(E_{real}) + (1-\log(1-E_{recon}))$;\newline
        \nl $D \ \longleftarrow D - \alpha (\partial \mathcal{L}_{D} / \partial D)$;\newline
        \nl $\mathcal{L}_{rec} \ \longleftarrow \vert \vert \ Y_i\ -\ \hat{Y_i}\ \vert \vert _{1}$;\newline
        \nl $\mathcal{L}_{id} \ \longleftarrow \vert \vert \ F_{vgg}(Y_i)\ -\ F_{vgg}(\hat{Y_i})\ \vert \vert _{1}$;\newline
        \nl $\mathcal{L}_{G} \ \longleftarrow \log(E_{real}) + \mathcal{L}_{rec} + \mathcal{L}_{id} $;\newline
        \nl $G \ \longleftarrow G - \alpha (\partial \mathcal{L}_{G} / \partial G);$
 } } 
 \caption{Our Learning Algorithm for the proposed model FEV-GAN}
 \label{ourAlgo}
\end{algorithm}

\section{Experiment}
To evaluate our model we performed an extensive experimental validation. In this following section, the experimental setup of our learning is detailed. Then, the model is evaluated quantitatively and qualitatively and compared to VGAN, MoCoGAN, ImaGINator and \citep{otberdout2019dynamic}. Finally, an ablation study is presented to observe the influence of each component of our model.

\subsection{Dataset}
The evaluation of our method is performed on:

\textbf{the MUG Facial Expression database (Multimedia Understanding Group Facial Expression database)} \citep{aifanti2010mug}: contains videos of $86$ people, performing seven facial expressions: "happiness", "sadness", "surprise", "anger", "disgust", "fear" and "neutral". Videos in the database start and end with neutral expressions and display the peak of the expression in the middle. Each video consists of $50$ to $160$ RGB frames of $896\times896$ resolution. The data of $52$ subjects is available to authorized Internet users, $25$ subjects are available on request and the remaining $9$ subjects are available only in the MUG laboratory. 
In this work we used the public data of $52$ subjects. We only used the first half of the six basic expressions videos, which starts with the neutral expression and ends with the expression peak.

\textbf{Oulu-CASIA NIR\&VIS facial expression database} \citep{zhao2011facial}: consists of 480 videos of $80$ people, between $23$ to $58$ years old, performing six facial expressions: "happiness", "sadness", "surprise", "anger", "disgust" and "fear". Each video consists of $9$ to $72$ RGB frames of $320\times240$ resolution, that begins with a neutral expression and end with the apex of the corresponding expression. The whole database is available to authorized Internet users.

\begin{table*}[ht]
\vspace{-4mm}
\caption{Quantitative comparison results of FEV-GAN and baseline models using PSNR, SSIM, ACD, and ACD-I metrics.}
\vspace{-4mm}
\begin{center}
\begin{tabular}{|c||c|c|c|c|||c|c|c|c|}
 \hline
          &  \multicolumn{3}{c}{Trained on MUG}  &&    \multicolumn{3}{c}{Trained on Oulu-Casia} & \\
    \hline
    \hline
     Model     &  PSNR     &    SSIM   & ACD & ACD-I  &  PSNR     &    SSIM   & ACD & ACD-I   \\
    \hline
    \hline
    VGAN \citep{vondrick2016generating} & $16.32$ & $0.41$ & $0.14$ & $1.55$ & $15.09$ & $0.61$ & $0.27$ & $1.37$ \\
  \hline
    C-VGAN \citep{vondrick2016generating} & $22.30$ & $0.83$ & $0.09$ & $0.73$ & $15.98 $ & $0.61$ & $0.25$ & $1.26$ \\
  \hline
  MoCoGAN \citep{tulyakov2018mocogan} & $18.16$ &  $0.58$ & $0.15$ & $0.90$ & $-$ &  $-$ & $-$ & $-$ \\
  \hline
  ImaGINator \citep{wang2020imaginator} & $20.29$ & $0.85$ & $0.08$ & $0.29$ & $22.98$ & $0.84$ & $0.07$ & $0.16$ \\
  \hline
  \citep{otberdout2019dynamic} & $25.9$ & $0.90$ & $-$ & $-$ & $24.44$ & $0.89$ & $-$ & $-$ \\
  \hline
  \textbf{The proposed FEV-GAN} & \textbf{27.10} & \textbf{0.91} & \textbf{0.09} & \textbf{0.23} & \textbf{25.61} & \textbf{0.89} & \textbf{0.12} & \textbf{0.19} \\
  \hline
 \end{tabular}
\end{center}
\label{resultTable}
\vspace{-5mm}
\end{table*}

\subsection{Implementation Details}
Before using the data, we first split the data in a subject independent manner into two sets, 80\% of the data for the learning and 20\% for the testing phases. We then cropped the face area and removed the background using OpenFace \citep{baltrusaitis2018openface}, normalized all videos to $32$ frames using the framework proposed in \citep{karcher1977riemannian}, and scaled each frame to $64\times64$ pixels. After the pre-processing phase, we ended up with $32\times64\times64\times3$ videos of different expressions and subjects with black background. 

The weights of the networks $E_{S}$, $ G_{dec} $ and $D$ are initially sampled from a Gaussian distribution of the mean zero and the standard deviation $0.01$. The weights of the network $E_{Id}$ are initialized with the weights of the pre-trained VGG-FACE \citep{parkhi2015deep} and frozen in the training phase. Gradient descent is used for $400$ epochs, in order to solve Eq \ref{eq:ModelOptimization}. Binary cross entropy loss is used for the adversarial loss, and $L_1$ norm is used for both reconstruction and identity losses. Additionally, Adam Optimizer \citep{kingma2014adam} is used to train the model with a learning rate of $0.0002$ and a Momentum of $0.5$. The pixels values of the input image and videos are scaled to the interval $[-1.1]$. Each layer of $G_{dec}$ is followed by the activation function ReLU \citep{agarap2018deep} and batch normalization \citep{ioffe2015batch}, except for the output, which use tanh. LeakyReLU \citep{xu2015empirical} and batch normalization are used in the discriminator except for the input layer. The implementation of the network is done with the TensorFlow Framework based on the implementation of \citep{vondrick2016generating}. The training on MUG data takes approximately 70 hours on an Nvidia GeForce GTX 1650 GPU (4Gb of memory). The training on Oulu-CASIA NIR\&VIS takes approximately 40 hours on an Nvidia Titan V GPU (12GB of memory). 

\subsection{Evaluation Metrics And Baselines}
To deeply evaluate our model quantitatively, we use several metrics: 

\textbf{1)} \textbf{PSNR}: (Peak Signal-to-Noise Ratio) measures the pixel-level similarity between the generated videos and their ground truth.

\textbf{2)} \textbf{SSIM}: (Structural Similarity Index Measure) represents the structural similarity between real and reconstructed videos.  

\textbf{3)} \textbf{ACD} \citep{tulyakov2018mocogan}: (Average Content Distance) measures content consistency in generated videos, based on the average of all pairwise $L_2$ distances between facial features of every two consecutive frames in a generated video. However, the ACD only represent identity consistency in the video, lacking information on the identity preservation of the input image.  

\textbf{4)} \textbf{ACD-I} \citep{zhao2018learning}: an ACD extension that measures the identity preservation of the input face in the generated video. It calculates the average of $L_2$ distances between the facial features of the frames of the video and the input image. To extract the facial feature vectors, we use OpenFace \citep{amos2016openface}, which is a deep model trained for facial recognition that can outperform human performance.\\

Higher SSIM and PSNR scores indicate better generated videos quality, lower ACD scores indicate similar faces in consecutive generated video frames, and lower ACD-I values indicate higher similarity between faces in the input images and the generated videos. \\

Regarding the state-of-the-art models used for comparison, we used the public codes of VGAN and ImaGINator provided by the authors with some minor changes. Since we deal with facial expression, we trained two versions of VGAN, the original version and a conditional version. In the conditional VGAN we used an encoder that transforms the input image to a latent code, the latent code is then concatenated with the labels and fed to the generator, that generates videos of the target expressions of the same input face. For MoCoGAN, we utilized the results given in \citep{wang2020imaginator}. As for \citep{otberdout2019dynamic}, we used the results in the original paper.
\begin{figure*}[ht]
\begin{center}
\includegraphics[scale=0.7]{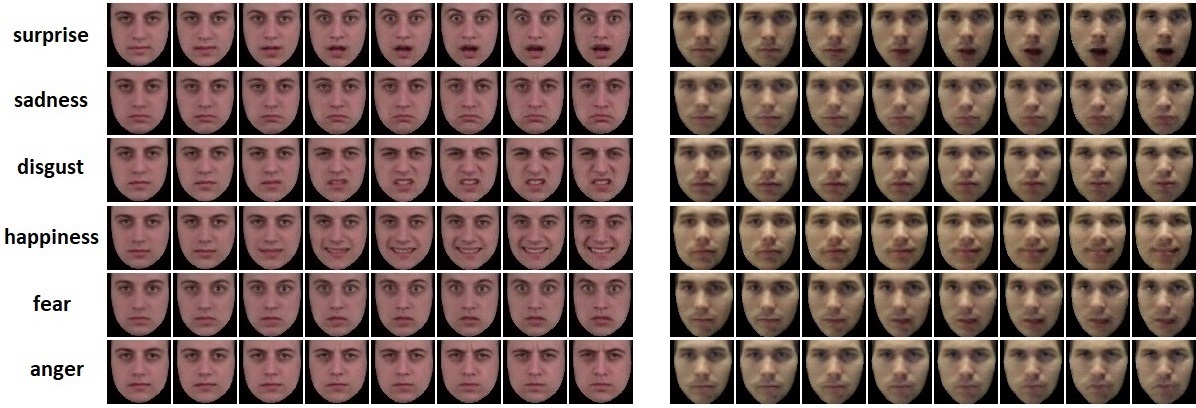}
\caption{The generation of facial expression videos on the MUG database (left) and Oulu-Casia (right). The image sequences show the six basic facial expressions of the same subject on the test dataset. The presented images are sampled every two frames. More examples of different identities are given in supplementary material.}
\label{ourResult}
\end{center}
\end{figure*}
\begin{figure*}[t] %float with two figures
\centering
\includegraphics[width=0.85\linewidth]{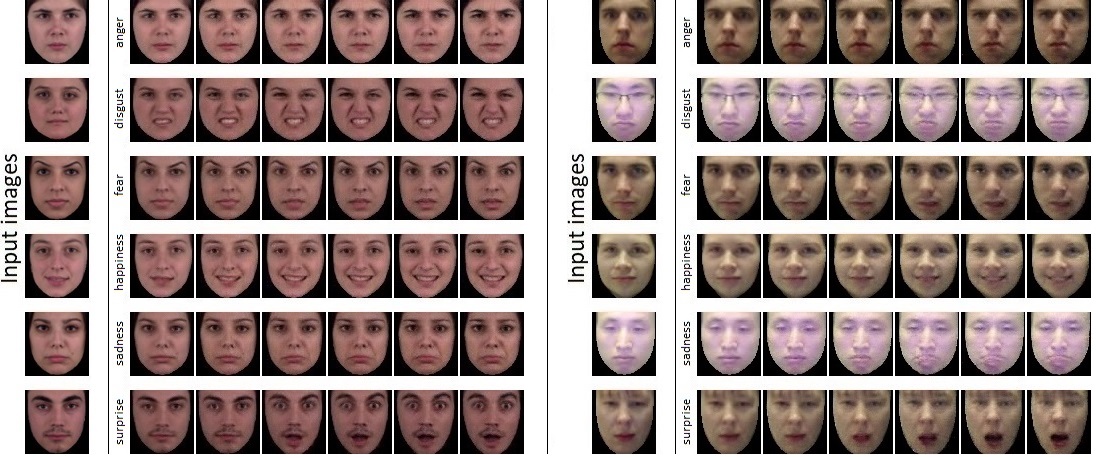}
\caption{Examples of videos generated by our model of the six basic facial expression performed by a person given in the input image.}
\label{examplesGenerated}
\end{figure*}
\begin{figure*}[ht]
\begin{center}
\includegraphics[scale=0.68]{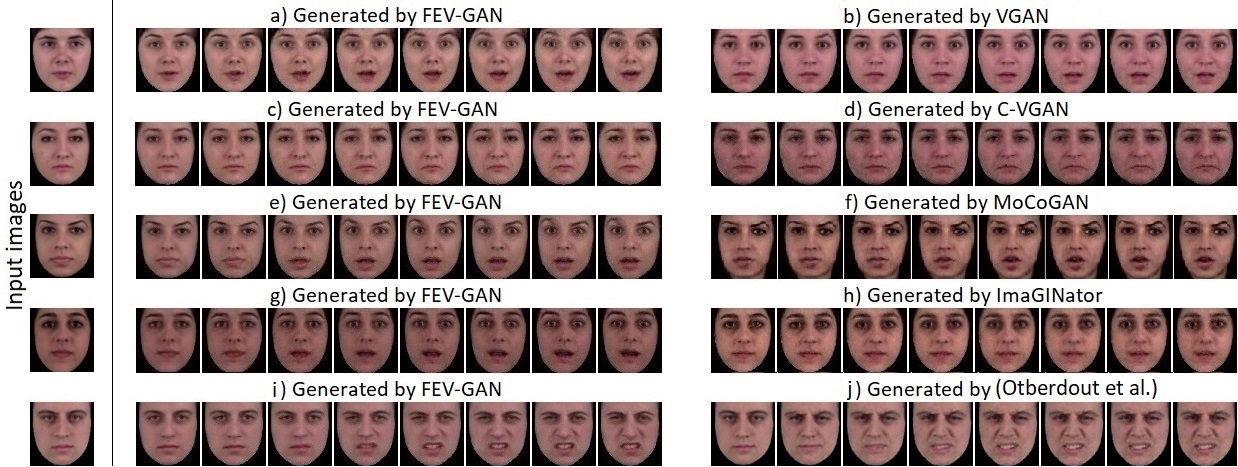}
\caption{Qualitative comparison of sequences generated by FEV-GAN model and by the state-of-the-art models on the MUG database.  The sequences generated by our model (a, c, e, g, i), by VGAN (b), C-VGAN (d) and by ImaGINator (h) are randomly selected from the test results. The sequence generated by MoCoGAN (f) and by \citep{otberdout2019dynamic} (j) are taken from the original papers. All the images are sampled with the time step of 4.  
}
\label{ourResult2}
\end{center}
\end{figure*}
\subsection{Experimental Results \& Analysis}

\textbf{Quantitative results -} to perform our quantitative analysis, we first generate $106$ videos of $18$ different subjects from the testing subset performing the six basic facial expressions.

First, we demonstrate that our model FEV-GAN offers better reconstruction capabilities than all baselines using \textbf{PSNR} and \textbf{SSIM}. Table \ref{resultTable} indicates that our model generates better quality videos with less noise, and preserves the general structure of the input through all the video. Then, we analyse the content consistency using \textbf{ACD} metric. Our model achieves similar content consistency to models with spatio-temporal convolutions (VGAN and ImaGINator) that are known to have high content consistency, while it surpasses MoCoGAN that uses Image-to-Image translation. As for identity preservation, our proposed model largely surpasses  VGAN, C-VGAN and MoCoGAN, and is competitive with ImaGINator.

We note that VGAN does not preserve the identity. Its score is used for the purpose of representing the metric values where identity is not preserved. The ACD and ACD-I comparison does not include the model used in \citep{otberdout2019dynamic} for the reason that the authors have used a different approach to calculate the metrics.

\begin{figure*}[ht]
\begin{center}
\includegraphics[scale=0.67]{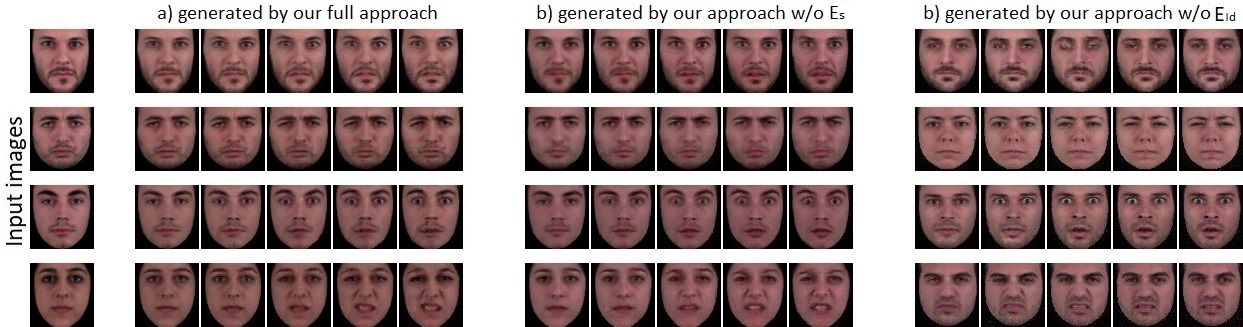}
\caption{Qualitative comparison of sequences generated by our full FEV-GAN model (a), by FEV-GAN w/o $E_{S}$ (b), and by FEV-GAN w/o $E_{Id}$ (c) on the MUG database.}
\label{ablationComparison}
\end{center}
\end{figure*}
\vspace{2mm}
\textbf{Qualitative results -} 
Fig.~\ref{ourResult} presents examples of generated videos of the six facial expressions of the same identity. Fig.~ \ref{examplesGenerated} demonstrate examples of generated videos of the six facial expressions of different given identities. Both figures show results of our proposed model trained on MUG dataset (left) and Oulu-Casia dataset (right). The generated videos are taken randomly from the test results. We recall that the data used for learning and testing the model is subject independent. The synthesized videos display generally natural, smooth and continuous facial variations that can be controlled according to the input label. They also preserve the features of the input images, such as the identity, the beard and the glasses. Our model shows slightly better results on MUG data than on Oulu-Casia data. This is due to the fact that Oulu-Casia contains fewer data instances with a variety of skin colors, lighting conditions and with accessories (glasses) on the face, which increases the difficulty of pattern learning for the model.
More examples of videos generated by our model are given in the Fig.~\ref{appendinx1} in Appendix.

In Fig.~\ref{ourResult2}, a comparison between our results and the state-of-the-art on the MUG database is conducted. The figure illustrates videos generated by our model, VGAN,  C-VGAN, MoCoGAN, ImaGINator and \citep{otberdout2019dynamic}. On the left side of the image, we show the input images utilized to generate the videos, in the middle, we show the videos generated by our model (a, c, e, g, i), and on the right side we show the videos generated by the baselines (b, d, f, h, j). In each comparison, each line, the videos generated by our model and the baseline contain the same identity and expression. In the first comparison (a,b), VGAN shows a natural smooth expression, but a total loss of the identity. In (c,d), C-VGAN generally preserves the input identity but with detail loss (like skin color), noise and artifacts. In (e,f), MoCoGAN offers sufficient identity preservation, but the identity is already used in the training set. MoCoGAN also suffers from the unnatural expression and distortions in the generated images, we suspect this causes the high ACD-I value even when the identity is preserved better than C-VGAN. In (g,h), ImaGINator preserves the input identity, but changes the skins color and displays an unnatural expression.
In (i,j), the baseline shows sufficient identity preservation and structure consistency, but displays artifacts near the mouth and nose area. The figure demonstrates that our model generally surpasses the baselines in terms of identity preservation and quality of the generated videos. It also demonstrate that our model maintains the same content-consistency and expression naturalness as spatio-temporal models. More qualitative comparison with these methods and others are conducted in Fig.~\ref{appendinx2} in Appendix.\\

In addition, we performed a subjective test, in which we aim to obtain opinions of human raters on videos generated by our method in comparison to the state-of-the-art. We asked $17$ volunteers to compare videos generated by our method and by the baselines. All the volunteers are PhD students and researchers in the deep learning field, from different university laboratories. The volunteers were asked more than 30 questions, where, each time we offered them two videos, generated by our model and by one of the baselines, without revealing the source of the videos. The question they were asked is to choose the best one of two given videos based on the average of expression naturalness, identity preservation and video quality. As shown in Table \ref{qualitativeResults}, our method largely outperforms the cited models. The raters show strong preference for our method over C-VGAN ($91.44\%$ VS $08.56\%$), MoCoGAN ($78.09\%$ VS $21.91\%$), ImaGINator ($71.07\%$ VS $28.93\%$), which corresponds to the quantitative results. The raters also commented that \textbf{1.} in comparison to C-VGAN, we show almost the same expression naturalness and content consistency, while surpassing it in identity preservation and quality of the generated videos. \textbf{2.} In comparison to MoCoGAN, we surpass it in all criteria. \textbf{3.} As for ImaGINator, we have the same level of identity preservation, but we offer better quality and more natural expressions. \textbf{4.} The raters also mentioned that the closest to our model is the one proposed in \citep{otberdout2019dynamic}. they have stated that the baseline slightly outperforms ours in identity preservation, while we generate videos with less noise and artifacts.
\begin{table}
\vspace{-4mm}
\caption{Subjective comparison of FEV-GAN and baseline models. The reported results are the mean of the raters preferences.}
\vspace{-7mm}
\begin{center}
\resizebox{1.2\textwidth}{!}{\begin{minipage}{\textwidth}
\begin{tabular}{|*{ 5}{c|}}
 \hline
     Models     &  Rater preference($\%$) \\
    \hline
    FEV-GAN/C-VGAN & $91.44\%$ / $08.56\%$ \\
  \hline
    FEV-GAN/MoCoGAN  & $78.09\%$ / $21.91\%$ \\
  \hline
    FEV-GAN/ImaGINator  & $71.07\%$ / $28.93\%$ \\
  \hline
   \begin{tabular}[x]{@{}c@{}}FEV-GAN\\/\citep{otberdout2019dynamic}\end{tabular}  & $68.05\%$ / $31.95\%$ \\
  \hline
 \end{tabular}
\end{minipage}}
\end{center}
\label{qualitativeResults}
\vspace{-9mm}
\end{table}

The results of the quantitative and the qualitative analysis prove that our model maintain the same content consistency and expressions naturalness as other spatio-temporal convolution models, while offering better quality and identity preservation.\\

\textbf{Ablation Study - }
In this section, we focus on demonstrating the importance of the techniques used to build our model. We showcase the effect of the double encoder method and the effect of the pre-trained facial feature extractor. This is conducted by performing the ablation study on the MUG database. We use the same evaluation metrics we used previously, on multiple versions of our model, where we cancel the target component and observe its effect. We first train two new other versions of our model. In the first one, we discard the spatial encoder $E_{S}$, while in the second one we remove the identity features encoder $E_{Id}$. Both networks are trained and evaluated using the same dataset, parameters, losses and for the same number of epochs, as the full network. Fig.~ \ref{ablationComparison} shows sequences generated by our full network, by the network w/o $E_S$, and by the network w/o $E_{Id}$. We deduce that our full network generates videos of the input identity performing the target expression with high quality and better facial details. As revealed in the sequences of Fig.~ \ref{ablationComparison}.\textbf{b}, the network w/o $E_S$ generates videos of lower quality and worse identity preservation, and it does not sufficiently preserve important details like the skin color, facial hair, and minor details like the surrounding areas of the mouth, nose and eyes.
As for Fig.\ref{ablationComparison}.\textbf{c}, the videos generated by network w/o $E_{Id}$ show more natural expressions but they also show more noise and distortions and they totally lack the identity preservation, .

The tests shown in Table \ref{ablationTable1} reveal that the full version of the network gives much better results than the modified versions in terms of quality, content consistency and identity preservation. This can be explained by the existence of the two encoders architecture. $E_{Id}$ capability to guarantee strongly identity related features makes it easier for $E_{S}$ to learn extracting other features that provide more details and information, giving the decoder the capability to learn generating better quality videos. 

\begin{table}[!ht]
\vspace{-1mm}
\caption{Performance comparison of our model without $E_{Id}$ and $E_{S}$}
\vspace{-7mm}
\begin{center}
\resizebox{1\textwidth}{!}{\begin{minipage}{\textwidth}
\begin{tabular}{|*{ 5}{c|}}
 \hline
     Model     &  PSNR     &    SSIM   & ACD & ACD-I   \\
  \hline
    FEV-GAN w/o $E_{S}$ & 25.44 & 0.90 & 0.11 & 0.34 \\
  \hline
    FEV-GAN w/o $E_{Id}$ & 19.71 & 0.69 & 0.13 & 1.22 \\
  \hline
    \textbf{Full FEV-GAN} & \textbf{27.10} & \textbf{0.91} & \textbf{0.09} & \textbf{0.23} \\
  \hline
 \end{tabular}
\end{minipage}}
\end{center}
\label{ablationTable1}
\vspace{-8mm}
\end{table}

\subsection{Discussion and Limitations}
\begin{figure*}[ht]
\begin{center}
\includegraphics[scale=0.68]{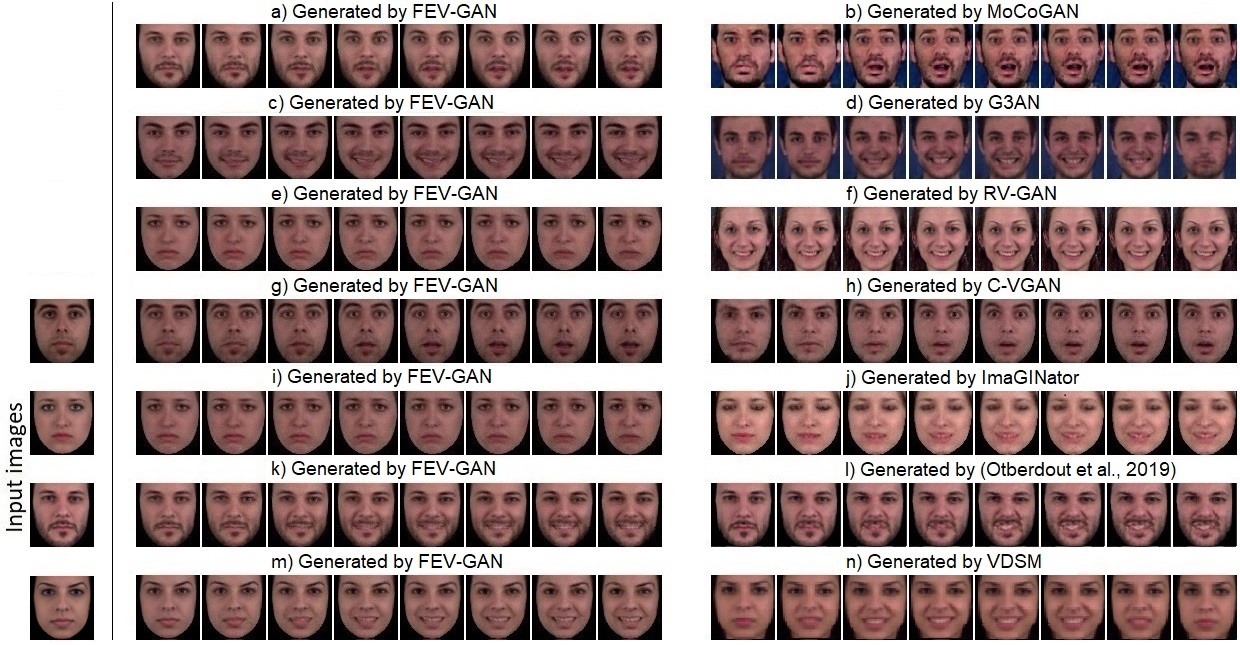}
\caption{Qualitative comparison of sequences generated by FEV-GAN model and by the state-of-the-art models on the MUG database.  The sequences generated by our model (a, c, e, g, i, k, l), by C-VGAN (h) and by ImaGINator (j) are randomly selected from the test results. The sequence generated by MoCoGAN (b), \citep{otberdout2019dynamic} (h), G3AN (d), RV-GAN (f), and by VDSM (n) are taken from the original papers. The absence of input identity indicate that the baseline does not preserve the input identity.}
\label{appendinx2}
\end{center}
\par  We notice MoCoGAN (b) shows large artifacts and generates videos where the intensity of the expression does not increase continuously. The video starts with neutral expression and suddenly turns to the target expression. G3AN (d) shows the transition between the expression, but the transition frames are noisy and contain artifacts. RV-GAN (f) video displays better quality and expression smoothness. However, these methods lack the identity preservation, which is very important when dealing with facial expression generation. C-VGAN (h) shows smooth facial expression and preserves the main structure of the input identity but loses much important details. ImaGINator (j) sufficiently preserves the identity but the expressions are unnatural and the generated frames are distorted.  \citep{otberdout2019dynamic} (l) and VDSM\citep{vowels2021vdsm} (n) show natural smooth facial expressions, and sufficient identity preservation. Nonetheless, VDSM images are blurry and \citep{otberdout2019dynamic} images display artifacts near the mouth area. Our generated samples present smoother and continuous facial expression evolution, sufficient identity details preservation, and they show less noise and artifacts compared to the stat-of-the-art models.
\end{figure*}

From the quantitative and qualitative comparisons, we conclude that our model largely outperforms the benchmarks C-VGAN and MoCoGAN in all criteria. As for the comparison to ImaGINator, we show similar content consistency, however ImaGINator results display some identity detail loss and unnatural expressions with clear distortions in the mouth and eyes area Fig.~\ref{ourResult2}\textbf{.f} ~Fig.~\ref{appendinx2}\textbf{.j} . For \citep{otberdout2019dynamic} , It generates videos with high identity and structure preservation, but it also shows artifacts in the mouth and eyes area . We argue that our proposed model generates generally better quality videos of natural expressions of the input identity, with minimum noise or distortions.

\begin{figure}[h] %float with two figures
\centering
\includegraphics[width=1\linewidth]{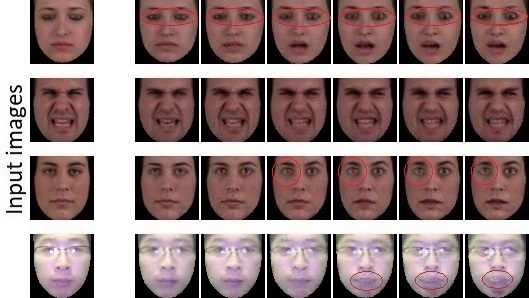}
\caption{Examples of flawed video generation by our model.}
\label{limitations}
\end{figure}

However, there are some inaccuracies that flaw our FEV-GAN model. Fig.~\ref{limitations} illustrates some examples of imprecise video generation. 
For example, the model learns to synthesis the teeth region when not given in the source image, but fails at generating the eyes when they are not clearly displayed in the input image ($1^{st}$ example of Fig.~\ref{limitations}). 
Also, the model is trained with neutral expression input images. If given a non-neutral expression, the generated video does not display the transition from neutral to the target expression ($2^{nd}$ example of Fig.~\ref{limitations}). 
In addition, the model occasionally constructs videos with some minor distortions in the eyes or mouth regions ($3^{rd}$ and $4^{th}$ example in Fig.~\ref{limitations}).

\section{Conclusion and perspectives}
In this paper, we present a novel Conditional GAN, namely FEV-GAN, for effectively generating the six basic facial expressions videos, given an input single neutral image and a target facial expression category. Specifically, we address the low quality and identity preservation issues encountered by facial expressions generation models that utilize fractionally strided spatio-temporal convolutions.\\

Based on our state-of-the-art study, these issues are related to the difficulty of the task of generating dynamic facial expressions. Our FEV-GAN model remedies these issues by utilizing two distinct encoders $E_{Id}$ (identity-encoder) and $E_S$ (spatial-encoder) that extract respectively the identity features $F_{Id}$ and spatial features $F_S$. These features are given as input to the decoder $G_{dec}$ in order to better preserve the identity and generate high quality facial expression videos.\\

We have deeply evaluated our method on two benchmark databases, The MUG facial expression database and Oulu-CASIA NIR\&VIS facial expression database, quantitatively by using different metrics (PSNR, SSIM, ACD, ACD-I), and qualitatively by using expert human eye rating. The results of these tests confirm our claims and show that our method significantly surpasses the state-of-the-art baselines in dynamic facial expression generation.\\

To further this research, we plan to test the model with other state-of-the-art facial recognition encoders, such as FaceNet \citep{DBLP:journals/corr/SchroffKP15}, OpenFace \citep{baltrusaitis2018openface}, and VGG-FACE2 \citep{DBLP:journals/corr/abs-1710-08092}. We also project to adapt our model to utilize facial landmarks sequences and/or facial action units instead of one-hot vector as a target category label. This will offer a large variation of possible expressions and more control over the generated expression.Another interesting perspective is to extend this approach to 3D facial expression video and 2D/3D human action generation.

\bibliographystyle{model2-names}
\bibliography{refs.bib}

\begin{figure*}[ht]
\begin{center}
\includegraphics[scale=0.8]{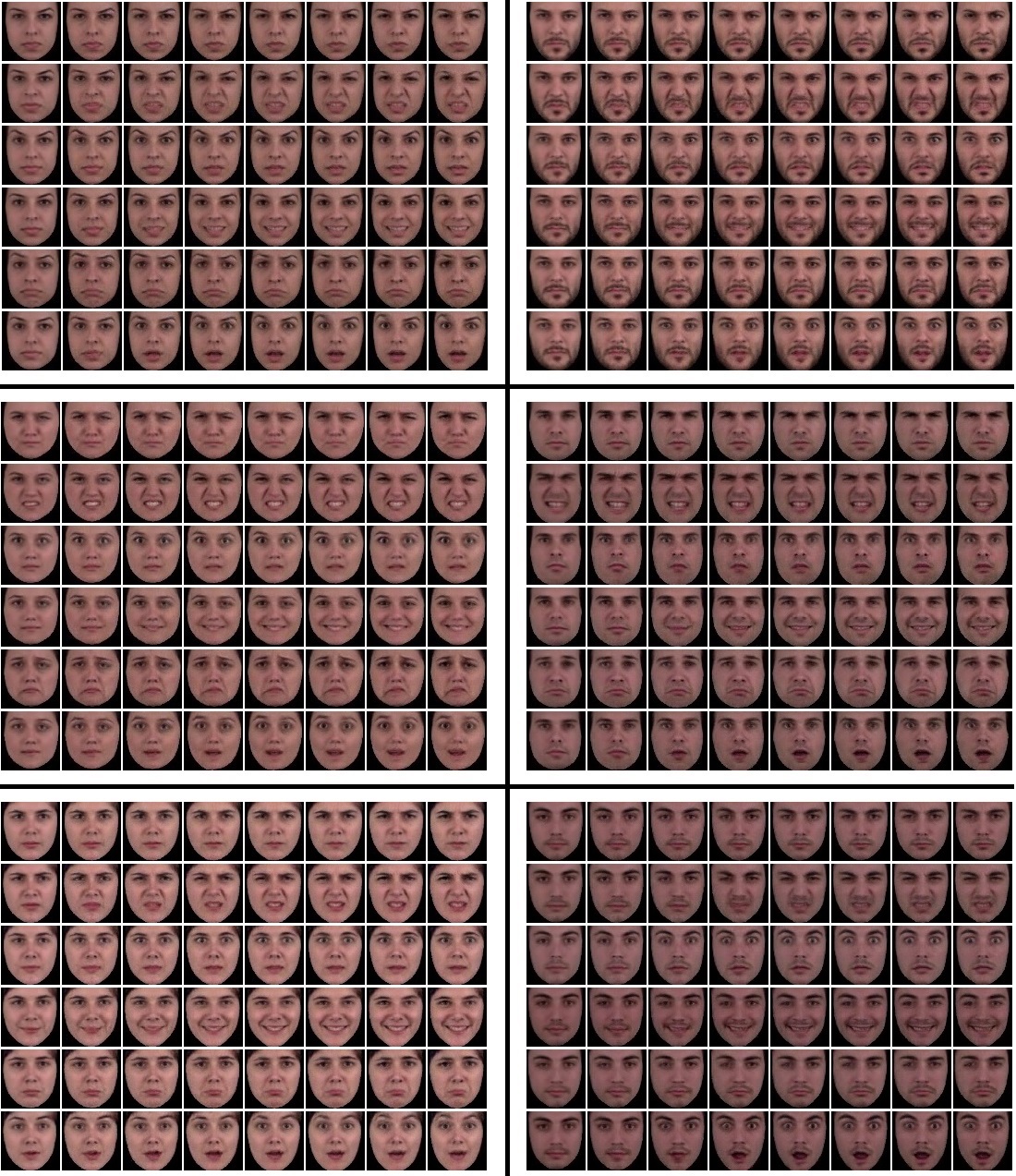}
\caption{examples of sequences generated by our proposed model. The image sequences in each box show the six basic facial expressions performed by the same subject on the MUG test dataset.}
\label{appendinx1}
\end{center}
\end{figure*}

\end{document}